\documentclass[10pt,twocolumn,letterpaper, twoside]{article}

\usepackage{wacv}
\usepackage{times}
\usepackage{epsfig}
\usepackage{graphicx}
\usepackage{amsmath}
\usepackage{amssymb}
\usepackage{booktabs}
\usepackage{amstext}
\usepackage{amsfonts}
\usepackage{multirow}
\usepackage{multicol}
\usepackage{caption}
\usepackage{subcaption}
\usepackage{xcolor}
\usepackage{breakcites}
\usepackage{fancyhdr}

\def\etal{et~al. } 
\def\vs{vs.~} 

\newcommand{\secref}[1]{Section~\ref{sec:#1}}

\newcommand{\tabref}[1]{Table~\ref{tab:#1}}

\def\wacvPaperID{1191} 

\wacvalgorithmstrack   

\wacvfinalcopy 

\ifwacvfinal
\usepackage[breaklinks=true,bookmarks=false]{hyperref}
\else
\usepackage[pagebackref=true,breaklinks=true,colorlinks,bookmarks=false]{hyperref}
\fi

\begin{document}

\title{PatchZero: Defending against Adversarial Patch Attacks by Detecting and Zeroing the Patch}

\author{Ke Xu$^*$ \and
Yao Xiao$^*$ \and
Zhaoheng Zheng \and
Kaijie Cai \and
Ram Nevatia \\
{\tt\small \{kxu47918, yxiao915, zhaoheng.zheng, kaijieca, nevatia\}@usc.edu}
}

\maketitle
\def\thefootnote{*}\footnotetext{Equal contributions}\def\thefootnote{\arabic{footnote}}


\begin{abstract}

Adversarial patch attacks mislead neural networks by injecting adversarial pixels within a local region. Patch attacks can be highly effective in a variety of tasks and physically realizable via attachment (e.g. a sticker) to the real-world objects. Despite the diversity in attack patterns, adversarial patches tend to be highly textured and different in appearance from natural images. We exploit this property and present PatchZero, a general defense pipeline against white-box adversarial patches without retraining the downstream classifier or detector. Specifically, our defense detects adversaries at the pixel-level and ``zeros out" the patch region by repainting with mean pixel values. We further design a two-stage adversarial training scheme to defend against the stronger adaptive attacks. PatchZero achieves SOTA defense performance on the image classification (ImageNet, RESISC45), object detection (PASCAL VOC), and video classification (UCF101) tasks with little degradation in benign performance. In addition, PatchZero transfers to different patch shapes and attack types. 

\end{abstract}

\section{Introduction}
\label{sec:intro}

\begin{figure}[!t]
    \centering
    \begin{subfigure}[b]{0.45\columnwidth}
         \centering
         \includegraphics[width=\columnwidth, height=0.75\columnwidth]{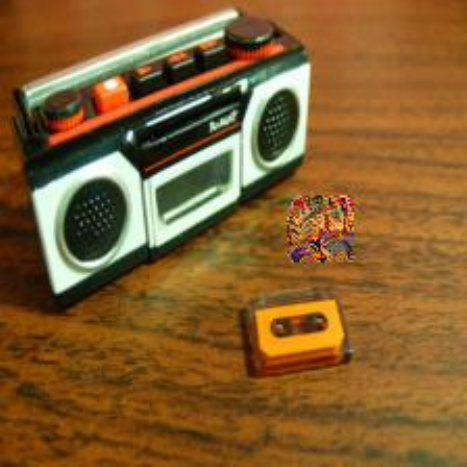}
         \caption{\textcolor{red}{\textsc{Parachute}}}
     \end{subfigure}
     \hspace{0.4em}
     \begin{subfigure}[b]{0.45\columnwidth}
         \centering
         \includegraphics[width=\columnwidth, height=0.75\columnwidth]{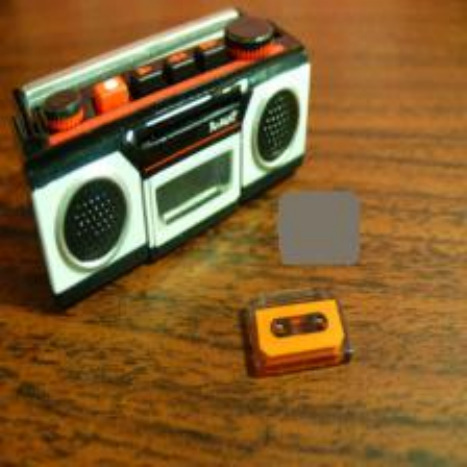}
         \caption{\textcolor{green}{\textsc{Cassette Player}}}
     \end{subfigure}

     \begin{subfigure}[b]{0.45\columnwidth}
         \centering
         \includegraphics[width=\columnwidth, height=0.75\columnwidth]{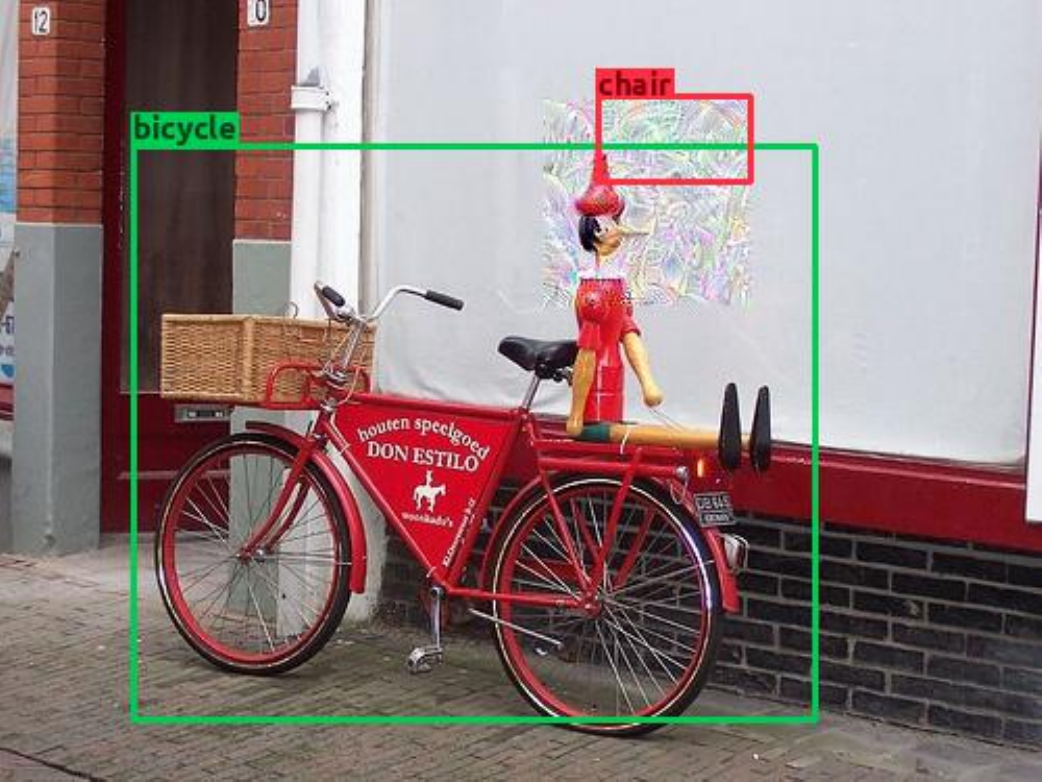}
         \caption{\textcolor{red}{\textsc{Chair}}, \textcolor{green}{\textsc{Bicycle}}}
     \end{subfigure}
    \hspace{0.4em}
    \begin{subfigure}[b]{0.45\columnwidth}
         \centering
         \includegraphics[width=\columnwidth, height=0.75\columnwidth]{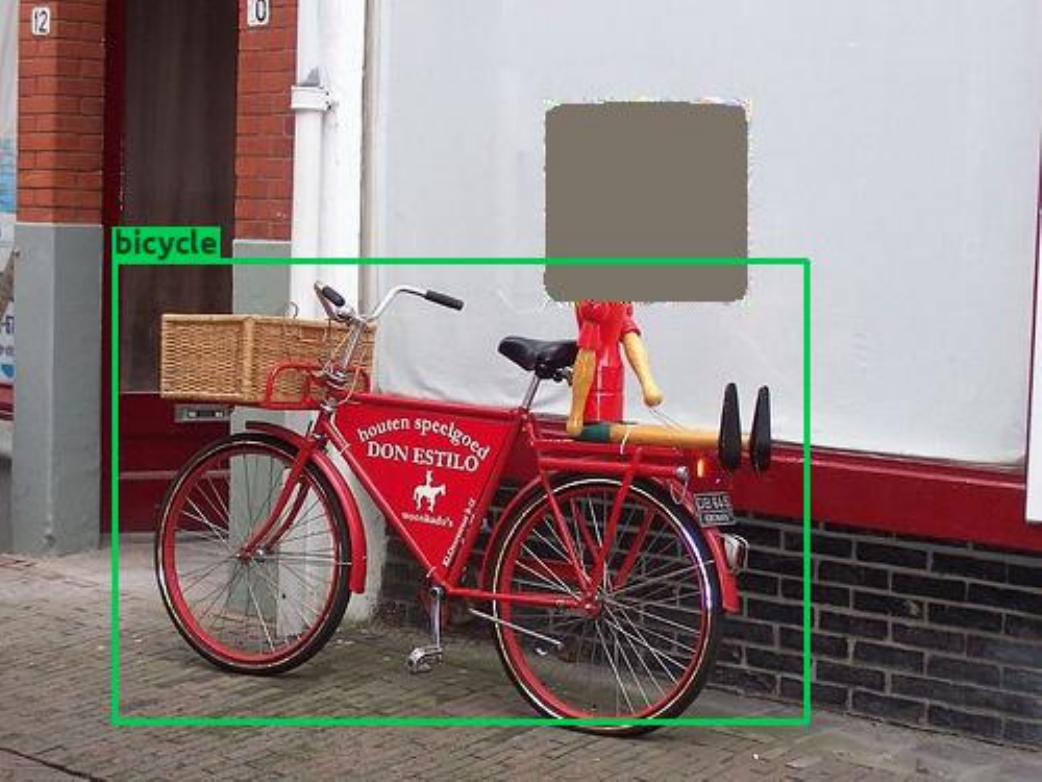}
         \caption{\textcolor{green}{\textsc{Bicycle}}}
     \end{subfigure}
     
     \begin{subfigure}[b]{0.45\columnwidth}
         \centering
         \includegraphics[width=\columnwidth, height=0.75\columnwidth]{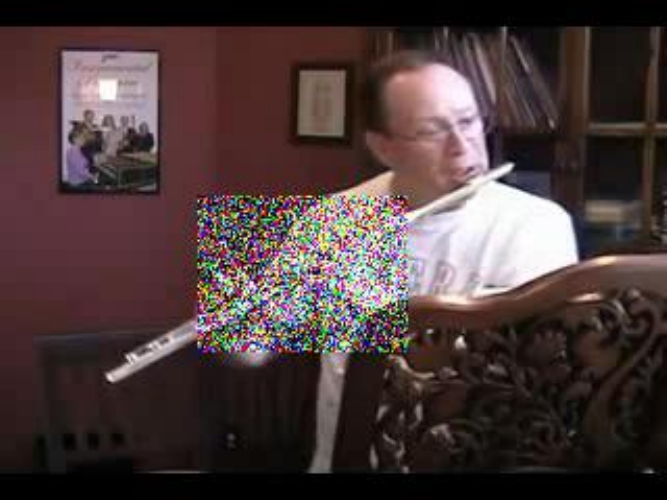}
         \caption{\textcolor{red}{\textsc{Drumming}}}
     \end{subfigure}
     \hspace{0.4em}
     \begin{subfigure}[b]{0.45\columnwidth}
         \centering
         \includegraphics[width=\columnwidth, height=0.75\columnwidth]{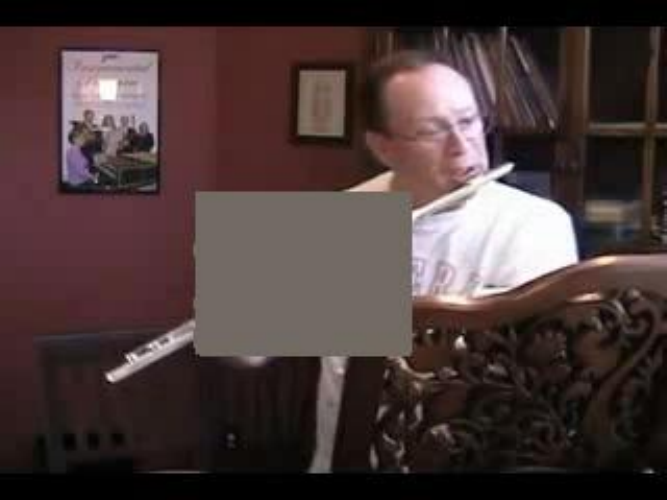}
         \caption{\textcolor{green}{\textsc{PlayingFlute}}}
    \end{subfigure}
    \caption{\textbf{PatchZero defense for adversarial patch attacks}. PatchZero takes an adversarial image (left) as input and outputs a processed image (right)  with adversarial pixels effectively removed. Our approach can be applied to the image classification (top), object detection (middle), and video classification (bottom) tasks without any retraining or modification of the downstream classifiers or detectors. Green and red denote correct and incorrect predictions.}
        \label{fig:teaser}
\end{figure}

 Early adversarial image attacks~\cite{goodfellow2014explaining,carlini2017towards,MadryMSTV18} perturb pixels over  the entire image; while these attacks are highly effective, they are hard to realize in the physical environment. This has led to study of  adversarial patch attacks that inject adversarial pixels within specified local regions, from the early Adversarial Patch~\cite{brown2017adversarial}, LaVAN~\cite{karmon2018lavan}, Masked Carlini-Wagner~\cite{carlini2017towards}, and Masked PGD~\cite{madry2019deep} attacks to the more recent DPatch~\cite{liu2018dpatch}, Robust DPatch~\cite{lee2019physical}, and Masked AutoPGD~\cite{croce2020reliable} attacks. 
 Patch attacks are physically realizable as they can be printed and placed into the scene. Real-world safety-critical computer vision systems, such as autonomous driving and security surveillance, are vulnerable to adversarial patches in the targeted scenes.

To tackle adversarial patch attacks in different domains, many patch defense methods have been proposed. Most of the defenses focus on image classification~\cite{hayes2018visible,naseer2019local,gowal2019scalable,levine2020randomized,chiang2020certified,zhang2020clipped,xiang2021patchguard,xiang2021patchguard++,xiang2021patchcleanser}, while object detection defenses~\cite{zhou2020information,chiang2020detection,xiang2021detectorguard} and video classification defenses~\cite{anand2020adversarial,lo2020defending,lo2021overcomplete} are relatively underexplored. Moreover, most of the defenses cannot be easily adapted to a different task and very few of them consider adaptive attacks. Some  ~\cite{xiang2021patchguard,xiang2021patchcleanser} also require prior knowledge, such as adversarial patch size, to be effective. In this study, we focus on adversarial patch attacks under the white-box setting, as they are stronger~\cite{andriushchenko2020square, wang2021di} than the black-box counterparts. We aim to design a general defense pipeline that can be easily applied to different classification and detection tasks under adaptive attacks, and does not require any prior attack knowledge. 

Our defense is based on the observation that although adversarial patches are localized, they can mislead predictions for objects far away in the image by exercising abnormally large influence on the spatial context due to their highly textured patterns. As shown in the first column of Figure~\ref{fig:teaser}, adversarial patches often have quite distinct texture and color distributions from those found in natural images. This observation leads to our idea to identify adversarial pixels with pixel-level patch detector and replace those pixels with mean pixel values (zero values after image normalization) to reduce or even eliminate their influences. Empirically, this process effectively ``zeros out'' the adversary and restores most of the accuracy for the downstream tasks. Therefore, we name our method ``PatchZero''. 

In an adaptive, white-box attack setting, the patch detector itself may be vulnerable to the attack. Since our patch detector outputs a binary mask which is non-differentiable during the backpropagation, we approximate the gradient of the binary mask using the Backward Pass Differential Approximation (BPDA)~\cite{athalye2018obfuscated} technique. 
We propose a two-stage adversarial training scheme to efficiently train PatchZero under BPDA attacks. 
The patch detector is first trained with DO attack examples and then reinforced with BPDA joint attack examples in successive stages. It is natural for the adversary and defender to train their models iteratively; however, it may be expected that this will always lead to a win for the adversary as the attacker gets to make the last call. One of our key contributions is in demonstrating that after some iterations of alternate training, the defense model becomes robust and able to detect the adaptive patches effectively.

We evaluate  PatchZero under the Masked PGD, Masked AutoPGD, and Masked CW attacks since they give a good coverage of the white-box patch attacks and can be easily applied to different tasks. PatchZero achieves state-of-the-art performances on all three tasks compared with the previous work, with little degradation in benign performance. Under the stronger BPDA adaptive attacks, the advantage margin of our defense method is even larger. Note that a recent paper ~\cite{liu2022segment} uses a similar defense approach for patch attacks on the object detection task. We are unable to compare with it directly as it uses different datasets and attack conditions. 

To summarize, our contributions are threefold: 
\begin{enumerate}
  \itemsep0em
  \item We present PatchZero, a general defense pipeline against white-box patch attacks that can be easily adapted to the tasks of image classification, object detection, and video classification without retraining of the downstream classifier or detector.
  \item We introduce a two-stage training scheme that reinforces PatchZero's robustness under the stronger adaptive attacks and accelerates training in the early stages.
  \item We evaluate our defense on multiple datasets and demonstrate generalization to different patch shapes and attack types.
\end{enumerate}

\section{Related Work}
\label{sec:rel}


\textbf{Patch Attacks:} Perturbation attacks manipulate the whole image to mislead neural networks. Patch Attacks, on the other hand, only modify a restricted region of the image. Brown \etal \cite{brown2017adversarial} first introduce the Adversarial Patch attack that generates a universal and physically realizable patch to mislead the image classification models. LaVAN~\cite{karmon2018lavan} is proposed at the same time but focuses on the digital patch. After the introduction of the full-image Carlini-Wagner (CW)~\cite{carlini2017towards}, PGD~\cite{madry2019deep} and AutoPGD~\cite{croce2020reliable} attacks, Masked CW, Masked PGD and Masked AutoPGD are three extensions to patch attacks by restricting the attack region.

We would like to mention some task-specific attacks. In the object detection domain, Liu \etal \cite{liu2018dpatch} design DPatch against popular object detectors. Lee \etal \cite{lee2019physical} investigate failure cases of DPatch and later introduce the Robust DPatch. Furthermore, Saha \etal \cite{saha2020role} introduce a blindness attack against the classifier inside an object detector, while Rao \etal \cite{rao2020adversarial} propose localization-optimized attacks. In the video classification domain, the only patch attack we can find besides Masked PGD and Masked AutoPGD attacks is the MultAV attack by Lo \etal \cite{lo2020multav}. MultAV is very similar to Masked PGD, but uses multiplication instead of summation when applying the perturbation.

We select Masked PGD and Masked AutoPGD for our experiments, since they can be easily applied across different tasks. Empirically, we also find them to be stronger than the task-specific attacks.

\textbf{Patch Defenses for Image Classification}: Digital Watermark(DW)~\cite{Hayes2018OnVA} and Local Gradient Smoothing (LGS)~\cite{naseer2019local} are among the early patch defenses. Both are later proved to be ineffective by Chiang \etal~\cite{chiang2020certified}, who proposes the first certified defense call Interval Bound Propagation (IBP). IBP limits the values of activation maps to guarantee a robustness lower bound. More recently, Xiang \etal~\cite{xiang2021patchguard} put forward PatchGuard, a network with small receptive field and outlier masking. It requires non-trivial modification of the backbone classifier. The same authors later propose another defense called PatchCleanser~\cite{xiang2021patchcleanser} that can be applied to any classifier. PatchCleanser uses an ensemble and exhaustive masking technique to identify the patch region. Both PatchGuard and PatchCleanser require the prior knowledge of the attack patch size to compute the optimal mask size and their certified robustness do not hold well for large patches. In comparison, our approach can defend against patch attacks of any patch sizes and shapes without any prior knowledge. 

\textbf{Patch Defenses for Object Detection}: Similar to the image classification defense,  object detection patch defense also receives a lot of attention recently. Liang \etal~\cite{liang2021catch} uses Grad-Cam to detect and filter out the unusual area of the image. However, Grad-Cam can only provide a coarse map and is subject to miss detection and false positives. Zhou \etal~\cite{9105983} combine Grad-Cam gradient map and discrete entropy to locate the adversarial pixels, but the detection results are still coarse and limited. DetectorGuard~\cite{xiang2021detectorguard} uses small receptive field CNN to output a robust objectness map that indicates the probability of objects being present at different locations. If the map results are different from the basic predictions, they will raise an alert for the adversary. However, this work only identifies but does not defend against patch attacks. SAC~\cite{liu2022segment} uses an approach of detecting and removing adversarial patches for the object detection task that is similar to ours. SAC uses identity mapping for binary mask gradient estimation, which we believe is weaker than our Sigmoid BPDA gradient estimation. It was proposed almost the same time as our approach and the code was not released, so we are not able to provide a direct comparison.

\textbf{Patch Defenses for Video Classification}: Adversarial patch defense is a relatively under explored research direction in video classification. Anand \etal~\cite{anand2020adversarial} propose Inpainting with Laplacian Prior (ILP) to detect and inpaint adversarial pixels in the Laplacian space. However, the method only works for optical-flow based video classifiers. Lo \etal propose to replace each Batch Normalization (BN)~\cite{ioffe2015batch} layer of a regular video classifier with three BNs. The network needs to be retrained adversarially to learn a ``switch mechanism'' to connect to the correct BN module. The same authors later propose OUDefend~\cite{lo2021overcomplete} module as an embedding feature denoiser to be inserted between the layers of a video classifier. Both methods from Lo \etal require modification and retraining of the downstream classifiers, while our approach can be plugged into any classifier.

\section{Defense Against Adversarial Patch Attacks}

In this section, we first introduce some related background in \secref{background}. Then we explain the PatchZero defense in \secref{pipeline}. Lastly, we elaborate on the two-stage training scheme for robustness against the stronger adaptive patch attacks in \secref{two-stage}.

\subsection{Background}
\label{sec:background}

\textbf{Projected Gradient Descent (PGD) and AutoPGD}: Introduced by Madry \etal \cite{MadryMSTV18}, PGD attack is one of the strongest perturbation attacks proven to be effective against image classification models. Given an input image $X$, its ground-truth label $Y^{*}$, model weights $\theta$ and the loss function $\ell$, PGD attack is generated by maximizing the loss function in an iterative manner: 

\begin{equation}
    X_{adv}^{(t+1)} = C_{\epsilon}\{X_{adv}^{(t)} + \alpha Sign(\nabla_{X} \ell(X_{adv}^{(t)}, Y^{*}, \theta)\}.
\label{eq:pgd}
\end{equation}

Note that the clipping function \textit{C} is utilized to prevent the per-pixel modification from going beyond the threshold $\epsilon$. In addition, random initialization and restarts are adopted to further strengthen the attack. AutoPGD~\cite{croce2020reliable} is later proposed as a PGD with auto step size tuning and a refined objective function. It is shown to be more effective than PGD under the same attack budget. 

\textbf{Masked PGD and Masked AutoPGD}: Although the original PGD attack is designed for the full-image perturbation attacks, it can be easily converted into a patch attack. As shown in Eq.~\ref{eq:masked-pgd}, only pixels inside the patch region $[x, y, h, w]$ will be modified by the PGD: 

\begin{equation}
\begin{split}
        & X_{adv}^{(t+1)}[patch] =  \\
        & \quad C_{\epsilon}\{X_{adv}^{(t)} + \alpha Sign(\nabla_{X} \ell(X_{adv}^{(t)}, Y^{*}, \theta)\}[patch].
\end{split}
    \label{eq:masked-pgd}
\end{equation}

Here \textit{patch} refers to the region defined as $[x:x+h, y:y+w]$ with the given $[x, y, h, w]$. Masked PGD can attack object detectors and video classifiers by deriving the gradients from corresponding loss functions. AutoPGD can be converted to its patch attack counterpart Masked AutoPGD in a similar manner.

\textbf{Adversarial Training:} Adversarial training \cite{goodfellow2014explaining,MadryMSTV18} has proven to be effective against various adversarial attacks. The key idea is to generate adversarial examples and inject them into the mini-batches during training. Generally, the effectiveness of adversarial training depends on the strength of adversarial examples. In practice, several researchers \cite{song2017pixeldefend,kannan2018adversarial,shafahi2020universal} have studied PGD attack and achieved significant robustness through adversarial training. To tackle adversarial patch attacks, we propose a two-stage training scheme that adversarially trains our models in two stages with samples produced by Masked PGD and Masked AutoPGD. The details are provided in Section~\ref{sec:two-stage}.

\subsection{PatchZero Network}
\label{sec:pipeline}

\begin{figure}[!t]
    \centering
    \includegraphics[width=0.98\columnwidth]{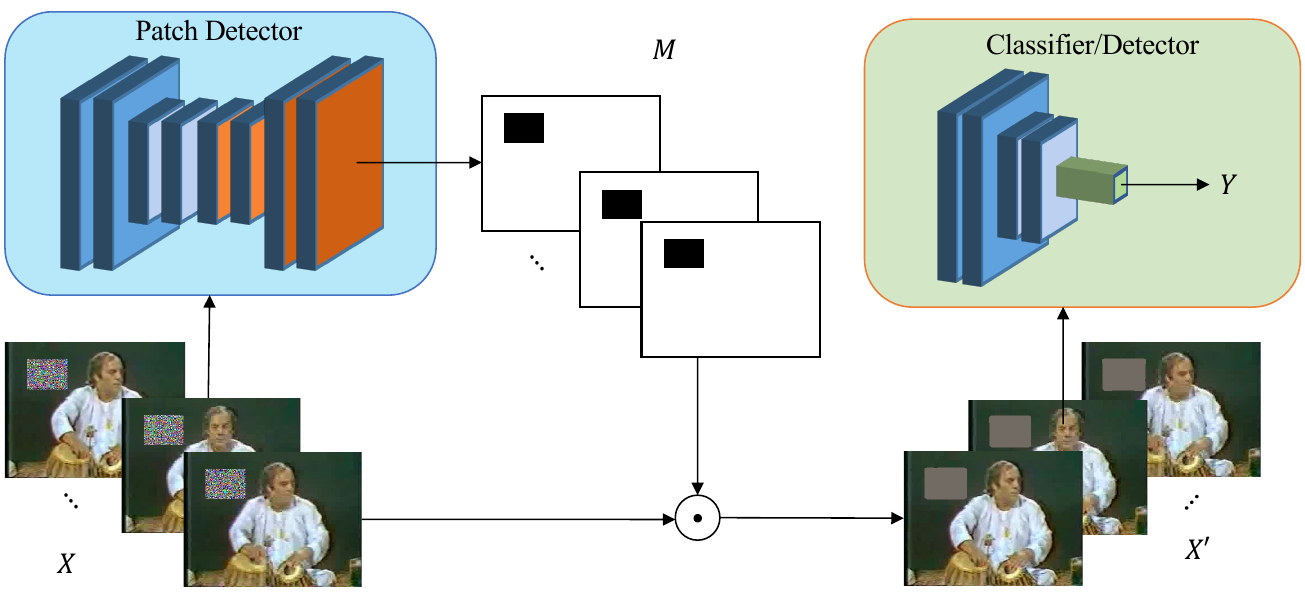}
    \caption{\textbf{Defense Pipeline of PatchZero.} The patch detector takes one or multiple attack images $X$ and predicts pixel-wise adversarial binary mask $M$ (black for adversarial pixels and white for benign pixels). We ``zero out'' the patch by multiplying $X$ with $M$ and fill the patch region with the mean pixel values. The preprocessed image $X'$ is passed to the downstream model for final predictions.}
    \label{fig:model}
\end{figure}

The full pipeline of PatchZero is shown in Figure~\ref{fig:model}. Our method consists of two steps. In the first step, the input image $X \in \mathbb{R}^{H\times W \times C}$ is processed by the patch detector $d: \mathbb{R}^{H\times W \times C} \rightarrow [0, 1]^{H \times W}$, which yields a probability map describing the possibilities for each pixel not being manipulated. We then binarize through a threshold $\epsilon_{p}$ the probability map to a binary mask $M \in \{0,1\}^{H\times W}$, where patch pixels are denoted zeros. For the second step, we remove the identified patch region via element-wise multiplication between $X$ and $M$. The masked region is then filled with mean pixel value $\overline{X}$ computed from the dataset to generate:
\begin{equation}
    X' = X \odot M + \overline{X} \odot \lnot M.
    \label{eq:neu}
\end{equation}
After the zero-out step, the downstream model $f$ takes the sanitized image $X'$ and makes the final prediction $Y$. 


We generate adversarial patches of random locations and sizes and the corresponding ground truth binary mask for each image. We construct the training set for $d$ by equally mixing attack images and benign images. During the patch detector training, we follow the loss function of PSPNet~\cite{zhao2017pyramid}, which consists of a main cross entropy loss and two auxiliary loss terms. During inference, the patch detector can detect and ``zero out'' the adversarial pixels most of the time, but occasionally misses some pixels at the border. To this end, we use morphological dilation to slightly enlarge the predicted mask by a few pixels.

\subsection{Adaptive Attack and Two-stage Training}
\label{sec:two-stage}

When generating adversarial patches, there are two attack strategies. In the Downstream-only (DO) attack, only gradients from the downstream classifier $f$ are considered. However, the patch detector $d$ itself is vulnerable to adversarial attacks, especially under the white-box setting, where attackers have full knowledge of the pipeline. In the stronger adaptive attack, both the gradients from the downstream classifier $f$ and from the patch detector $d$ are considered. As defined in Eq.~\ref{eq:neu}, the zero-out step includes the non-differentiable binarization operation. The pixel-level gradient $\nabla_{X}\ell(X_{adv}^{(t)}, Y^{*}, \theta)$ cannot be computed directly through back-propagation.

\textbf{BPDA Adaptive Attack:} Proposed by Athalye \etal~\cite{athalye2018obfuscated}, BPDA is an approximation strategy to bypass non-differentiable layers inside a network and achieve effective adaptive attacks. Given a non-differentiable operation $h$, BPDA finds a differentiable approximation $h'$ that satisfies $h(x) \approx h'(x)$. The original operation $h$ is used in the forward pass but replaced by the approximation $h'$ in the backward pass. To apply BPDA to PatchZero, we approximate the binarization operation by the Sigmoid function, since the binarization is essentially a Step function. With this approximation, we can utilize gradients from both the patch detector and the victim classifier or detector to generate attacks.

\textbf{Two-stage Adversarial Training:} The BPDA adaptive attack introduces some difficulty to the training process of the patch detector. In the early stages, the patch detector is immature and creates random gradients. Since the adaptive attack passes the gradients from the downstream classifier through the patch detector, the resulting gradients will be misleading. To resolve this issue, we propose a two-stage training scheme as described below:

\begin{itemize}
    \itemsep0em
    \item \textbf{Training Stage 1:} We first generate adversarial patches using the DO attack, which only consider the gradients from the downstream classifier or detector (Figure~\ref{fig:model} green box). We train the patch detector $d$ with a mixture of benign and adversarial images. 
    \item \textbf{Training Stage 2:} When the patch detector starts to converge on the DO attack images, we switch to the 2nd stage of training. We generate adversarial patches using the BPDA adaptive attack, which considers gradients from both parts (blue and green box of Figure~\ref{fig:model}) of the pipeline. We generate online adversarial attacks at every training step with updated model weights. This practice creates an attacker-defender race and further fortifies the effectiveness of the patch detector.
\end{itemize}

The two-stage training mechanism greatly accelerates the training process and improves the robustness of PatchZero under the stronger BPDA adaptive attacks.

\section{Experiments}
\label{sec:exp}

We adopt the PSPNet~\cite{zhao2017pyramid} with the ResNet-50~\cite{he2016deep} backbone as the patch detector of PatchZero. We initialize the PSPNet with weights pre-trained on the ImageNet\cite{deng2009imagenet} and follow the loss function for image segmentation. We train our PSPNet patch detector through the two-stage adversarial training introduced in \secref{two-stage}. Regarding the binarization threshold, we set $\epsilon_p$ as 0.5. We developed the PatchZero and the two-stage training scheme in PyTorch~\cite{paszke2019pytorch} and use the Adversarial Robustness Toolbox (ART)\footnote{https://github.com/Trusted-AI/adversarial-robustness-toolbox} for generating attacks. For the two-stage training, we use a learning rate of 0.0001, the Adam~\cite{kingma2014adam} optimizer, and a batch size of 64 for image classification, 16 for object detection, and 36 for video classification.

\subsection{Image Classification}
\label{sec:class-setup}

\begin{table}[!t]
    \begin{center}
    \begin{tabular}{@{}lcccc@{}}
    \toprule
    \textbf{Defense} & \textbf{Benign} & \textbf{MPGD} & \textbf{MAPGD} & \textbf{MCW} \\
    \cmidrule{1-5}
    \textbf{Undefended} & 81.62\% & 14.35\%  & 9.40\% & 49.57\% \\
    \textbf{GT Mask}  & 81.60\%    & 81.42\% & 81.34\% & 81.37\% \\
    \cmidrule{1-5}
    \textbf{PG~\cite{xiang2021patchguard}} & 60.40\% &  49.41\% &  48.91\%  &  56.95\%\\
    \textbf{PC~\cite{xiang2021patchcleanser}} & 80.54\% & 64.30\% & 63.57\% & 73.12\% \\
    \textbf{PZ (DO)} & \textbf{81.47\%} & \textbf{75.60\%} & \textbf{76.80\%} & \textbf{74.24\%} \\
    \cmidrule{1-5}
    \textbf{PZ (BPDA)} & 81.48\% &  55.46\% &  70.02\% & - \\
    \bottomrule
    \end{tabular}
    \end{center}
    \caption{\textbf{Benign and robust accuracy on the ImageNet classification dataset.} PG, PC, and PZ stand for PatchGuard, PatchCleanser, and PatchZero respectively.}
    \label{tab:image-baseline}
\end{table}

\noindent \textbf{Implementation Details:} 
We conduct our image classification experiments on two datasets. We use the validation split of the ImageNet~\cite{deng2009imagenet} dataset with 50,000 images and 1000 classes. We also evaluate on the RESISC-45~\cite{cheng2017remote} remote sensing dataset that contains 31,500 images and 45 scene classes. Compared with the ImageNet, RESISC-45 has a larger image size (256x256) and provides a remote sensing perspective. On the ImageNet, we use ResNet50-v2 as the backbone image classifier for all the defense methods. On the RESISC-45, we use DenseNet121~\cite{huang2017densely} as the image classifier. We use the top1 accuracy for evaluation. 

\noindent \textbf{Attacks:} 
For the Masked PGD (MPGD) attack, we use a perturbation strength of 1.0, a step size of 0.01 and 100 iterations. For the Masked AutoPGD (MAPGD) attack, we use a perturbation strength of 0.3, a step size of 0.1 and 100 iterations. For the Masked Carlini-Wagner (MCW) attack, we use a perturbation confidence of 0.5, a learning rate of 0.1 and 100 iterations. Following the same settings as the previous works, we use 2\% rectangular patches for ImageNet and 9\% square patches for RESISC-45. The patch sizes are w.r.t. the image area and patch locations are random.

\noindent \textbf{Baseline Defenses:}
\begin{itemize}
\itemsep0em
    \item \textbf{PatchGuard}: PatchGuard~\cite{xiang2021patchguard} is a certified defense with small receptive field and outlier masking. We empirically evaluate the robustness under the same attack settings. Note that the method requires  prior knowledge of the attack patch size to estimate the defense mask window size.     
    \item \textbf{PatchCleanser}: PatchCleanser~\cite{xiang2021patchcleanser} is another certified defense against adversarial patches via two rounds of exhaustive masking and ensemble. We empirically evaluate the robustness under the same attack settings. This defense method also requires prior knowledge of attack patch size.
    \item \textbf{JPEG Compression}: Guo \etal~\cite{guo2017countering} propose to defend against adversarial attacks through image transformations, including  JPEG compression. Here we use JPEG compression as a preprocessor defense.
    \item \textbf{Adversarial Training}: For each downstream model $f$, we follow a typical adversarial training scheme~\cite{MadryMSTV18} and train the downstream classifier with a mixture of clean and adversarial images.
    
\end{itemize}

\begin{table}[!t]
    \centering
    \begin{tabular}{@{}lccc@{}}
    \toprule
    \textbf{Defense} & \textbf{Benign} & \textbf{MPGD} & \textbf{MAPGD} \\
    \cmidrule{1-4}
    \textbf{Undefended} & 92.9\%  & 3.0\%   & 1.7\%   \\
    \textbf{GT Mask} & 92.9\%   & 87.8\%  &  87.2\%   \\
    \cmidrule{1-4}
    \textbf{JPEG Comp~\cite{dziugaite2016study}}      &  91.0\%   &  4.1\%  &  1.7\%\\
    \textbf{Adv Training~\cite{goodfellow2015explaining}}      &  83.9\%   &  71.8\%  &  67.2\% \\
    \textbf{PZ (DO)}     & \textbf{92.9\%}   & \textbf{87.5\%}  & \textbf{85.0\%}    \\ 
    \cmidrule{1-4}
    \textbf{PZ (BPDA)}   &  92.9\%   &  81.2\%  &  76.4\% \\
    \bottomrule
    \end{tabular}
    \caption{\textbf{Benign and robust accuracy on the RESISC-45 classification dataset.}}
    \label{tab:image-acc}
\end{table}

\noindent \textbf{Defense Results:}
We first present the undefended baseline and GT Mask baseline which assume perfect adversarial patch detection. As shown in \tabref{image-baseline}, the GT baseline recovers most of the robustness accuracy compared with no attack, showing the potential of our approach. The two certified defense baselines, PatchGuard and PatchCleanser, both require prior knowledge of attack patch size and the robustness declines as the patch size increases. Our method's performance is not strongly dependent on the patch size (except for occlusion effects) but we tested with  2\% patch size for fair comparison.

PatchZero outperforms PatchGuard by ~26\% and PatchCleanser by ~13\% on both the MPGD and MAPGD attacks. PatchZero has similar performance as PatchCleanser and both outperform PatchGuard by ~17\% on the  MCW attack. Compared with the GT Mask results, PatchZero has almost no drop in accuracy in all attacks except for MCW under the DO attack, but larger gaps under the stronger BPDA adaptive attack. Note that neither PatchGuard nor PatchCleanser can be easily adapted for adaptive attack.

\begin{table*}[!h]
    \centering
    \begin{tabular}{@{}lcccccc@{}}
    \toprule
    \multirow{2}{*}[-0.5em]{\textbf{Defense}} & \multicolumn{3}{c}{\textbf{Benign}} & \multicolumn{3}{c}{\textbf{MPGD}} \\ \cmidrule(lr){2-4}  \cmidrule(lr){5-7} & \textbf{AP}     & \textbf{AP50}    & \textbf{AP75}       & \textbf{AP}     & \textbf{AP50}    & \textbf{AP75}  \\ \cmidrule{1-7}
    \textbf{Undefended} & 49.20\%  & 76.4\%   & 52.6\%  & 6.5\% & 10.9\%  & 6.7\% \\
    \textbf{GT Mask} & 49.2\%  & 76.4\%  &  52.6\%  & 43.0 \% & 68.8\% & 44.4\%       \\
    \cmidrule{1-7}
    \textbf{JPEG Comp~\cite{dziugaite2016study}}      &  47.7\%   &  75.0\%  &  51.4\% &  30.0\% &  48.1\% &  32.3\% \\
    \textbf{Adv Training~\cite{goodfellow2015explaining}}      &  47.7\%   &  75.1\%  &  51.7\% &  16.8\% &  31.9\% &  15.2\% \\
    \textbf{PZ (DO)}  & \textbf{48.4\%}   & \textbf{75.3\%}  & \textbf{51.8\%}     & \textbf{41.5\%}     & \textbf{66.1\%}      & \textbf{43.8\%} \\ 
    \cmidrule{1-7}
    \textbf{PZ (BPDA)}           &  48.4\%   &  75.3\%  &  51.8\% &  35.1\% & 60.0 \% & 35.5 \% \\
    \bottomrule
    \end{tabular}
    \caption{\textbf{Benign and robust AP on the PASCAL VOC object detection dataset.}}
    \label{tab:VOC}
\end{table*}

\begin{table*}[!h]
    \centering
    \begin{tabular}{@{}lccccc@{}}
    \toprule
    \multirow{2}{*}[-0.5em]{\textbf{Defense}} & \multirow{2}{*}[-0.5em]{\textbf{Benign}} & \multicolumn{2}{c}{\textbf{MPGD}} & \multicolumn{2}{c}{\textbf{MAPGD}} \\ 
    \cmidrule(lr){3-4} \cmidrule(lr){5-6} & & \textbf{5\%} & \textbf{10\%} & \textbf{5\%} & \textbf{10\%} \\ 
    \cmidrule{1-6}
    \textbf{Undefended} & 94.55\% & 8.42\% & 3.96\% & 18.81\% & 0.00\% \\
    \textbf{GT Mask} & {94.55\%} & {91.58\%} & {93.07\%} & {91.58\%} & {93.07\%} \\
    \cmidrule{1-6}
    \textbf{Video Comp~\cite{halbach2003h}} & \textbf{94.55\%} & 21.29\% & 6.44\% & 12.87\% & 0.99\% \\
    \textbf{PZ (BPDA)} & \textbf{94.55\%} & \textbf{81.68\%} & \textbf{82.67\%} & \textbf{73.27\%} & \textbf{76.24\%} \\
    \bottomrule
    \end{tabular}
    \caption{\textbf{Benign and robust accuracy on the UCF101 video classification dataset.}}
    \label{tab:video-results}
\end{table*}

We also evaluated PatchZero on the RESISC-45 dataset to test robustness under higher image resolution and much larger patch sizes (9\% of image size), as shown in Table~\ref{tab:image-acc}. We compare with JPEG compression and adversarial training baselines. JPEG compression performs poorly; adversarial training shows much better defense but PatchZero performs better by a large margin, even under the stronger BPDA adaptive attack. Also, adversarial training reduces benign accuracy substantially (by 9\%), while PatchZero maintains the benign accuracy of the undefended model. 

For both datasets, we can see that the BPDA accuracy drops from GT accuracy, more seriously in ImageNet than in RESISC-45, likely due to much larger variety in the former. Nonetheless, substantial improvements are achieved over undefended model and available alternatives. Further improvements in the patch detection performance will be a consideration in our future research.

\subsection{Object Detection}
\label{sec:detection-setup}

\noindent \textbf{Implementation Details:} For the object detection task, we evaluate on the PASCAL VOC~\cite{everingham2010pascal} dataset, which has 20 object categories. Following the same setting as previous works~\cite{wu2019detectron2}, our models are trained on VOC 2007 plus VOC 2012 and tested on VOC 2007. We use the Faster-RCNN~\cite{ren2015faster} with ResNet-50~\cite{he2016deep} as the downstream detector. For evaluation, we use the standard Average Precision (AP), AP50, and AP75 metrics.

\noindent \textbf{Attacks:} We defend against Masked PGD attack with a perturbation strength of 0.3, step size of 0.1, and 100 iterations. Patch sizes are $120\times120$ and patch locations are random.

\noindent\textbf{Baseline Defense:} We adopt Adversarial Training, JPEG Compression as the baseline defense methods, since we are unable to find any other patch defense baselines and the two image classification baselines do not obviously transfer to the detection task.

\noindent \textbf{Defense Results:}
Table~\ref{tab:VOC} shows evaluation results against the Masked PGD attack on PASCAL VOC. Similarly, the GT Mask baseline assumes perfect patch detection and recovers most of the accuracy compared with no attack. PatchZero achieves an AP of 41.5\%, around 8\% lower than the benign performance, while JPEG compression and adversarial training only get 30.0\% and 16.8\% AP, respectively. The DO attack results of PatchZero are very close to the GT Mask results. PatchZero also outperforms the other baselines on the benign images. The BPDA results are lower compared with our DO results, but still 5\% higher than JPEG Compression and 18\% higher than Adversarial Training, even though they use the much weaker DO attack. 

\subsection{Video Classification}
\label{sec:videoclassify}

\noindent \textbf{Implementation Details:} We conduct our video classification experiments on the UCF101, an action recognition dataset that has 13,320 short trimmed videos from 101 action categories. Since adversarial defense is computationally expensive on the video domain, we randomly select 202 video from the test dataset. We adopt the MARS~\cite{huang2017densely} model as the downstream classifier. We use the top\@1 and top\@5 classification accuracy as the evaluation metrics.

\noindent \textbf{Attacks:} For video classification, we consider the Masked PGD and Masked AutoPGD attacks, with perturbation strength of 1.0, step size of 0.2, and 20 iterations. All attacks use BPDA and have patch sizes of 5\% and 10\%. The patch locations are fixed for all frames of the same video but random for each video.

\noindent \textbf{Baseline Defenses:} Due to the lack of reliable adversarial patch defense methods in video classification, we pick the H.264 video compression~\cite{halbach2003h} as the baseline defense.

\noindent \textbf{Defense Results:}
We compare the defense performance of different defense methods in \tabref{video-results}. All methods use the MARS model as the downstream classifier. The GT baseline assumes perfect patch detection and recovers most of the robustness accuracy compared with no attack. For the benign videos, neither PatchZero nor the video compression degrade accuracy compared with the undefended MARS classifier. For the attack scenarios, Masked AutoPGD attack is stronger than the Masked PGD attack and attacks with larger patch sizes (10\%) are stronger. PatchZero significantly outperforms the video compression baseline under all attack combinations. The margin is even larger for the stronger Masked AutoPGD attacks and larger patch sizes. For example, for the 10\% Masked AutoPGD attacks, our method outperforms the Video Compression baseline by a margin of 75.25\% on the top1 accuracy. Compared with the GT Mask, PatchZero still has some performance gap, but already outperforms the other baseline by a large margin.

Lo \etal proposed ``3-BN''~\cite{ioffe2015batch} and ``OUDefend''~\cite{lo2021overcomplete} modules as defense for multiple video attacks. The authors do not provide implementation for either method, so we cannot thoroughly compare with them. In their only patch attack experiment, they use a much weaker, Downstream-only Masked PGD attack with patch size of 1.2\% , perturbation strength of 1.0, and 5 iterations.  The ``3-BN'' achieves a 63.8\% accuracy and OUDefend achieves a 42.00\% accuracy. In comparison, we use the adaptive version of Masked PGD attack and with stronger attack parameters: patch sizes of 5\% and 10\%, perturbation strength of 1.0, and 20 iterations. PatchZero achieves 81.68\% top1 accuracy, almost 20\% higher. Also, both methods require modification and adversarial training of the downstream video classifier. Neither of them can be easily applied across tasks. 

\begin{figure}[!t]
    \centering
    \captionsetup{type=figure}
    \begin{subfigure}[b]{0.3\columnwidth}
         \centering
         \includegraphics[width=\textwidth]{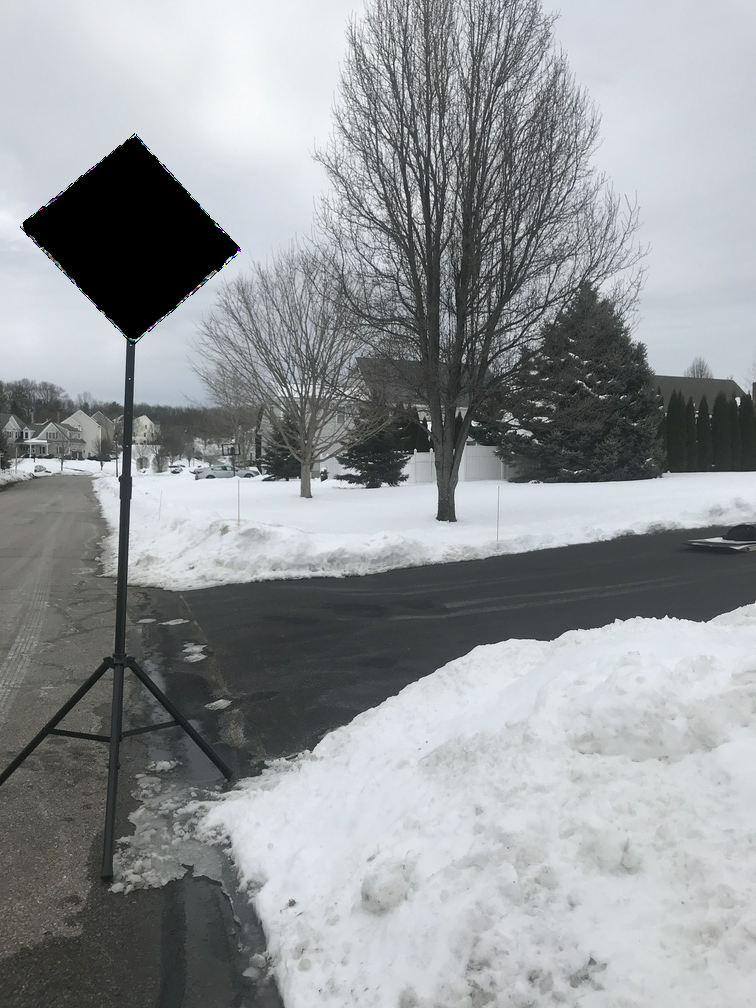}
         \caption{Diamond}
     \end{subfigure}
     \hspace{0.2em}
     \begin{subfigure}[b]{0.3\columnwidth}
         \centering
         \includegraphics[width=\textwidth]{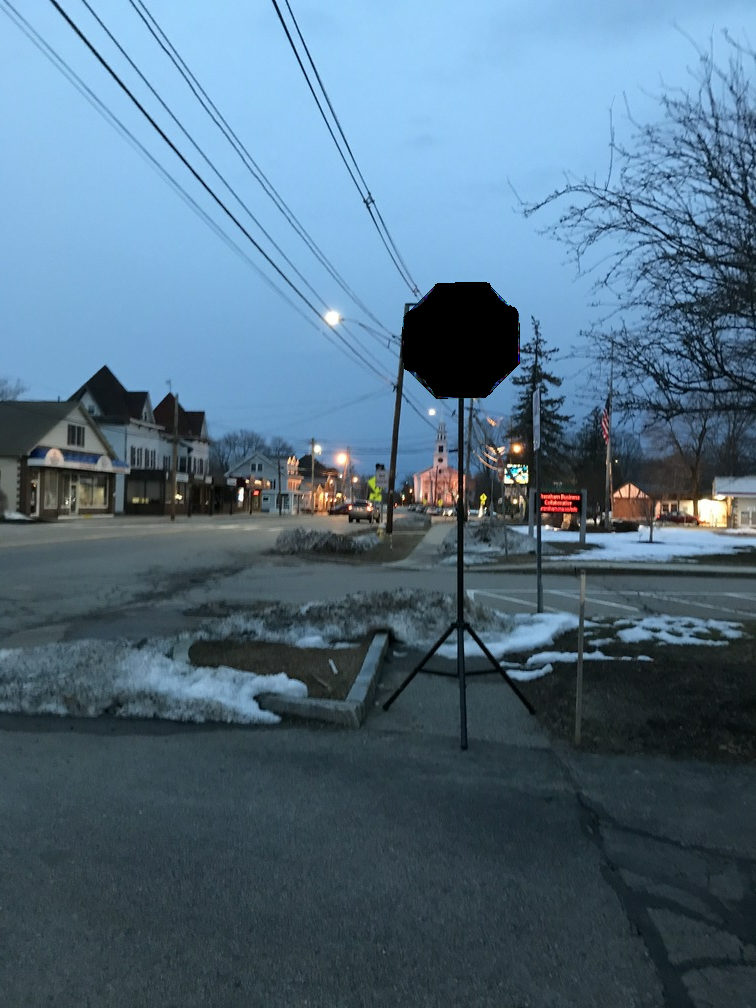}
         \caption{Octagon}
     \end{subfigure}
     \hspace{0.2em}
     \begin{subfigure}[b]{0.3\columnwidth}
         \centering
         \includegraphics[width=\textwidth]{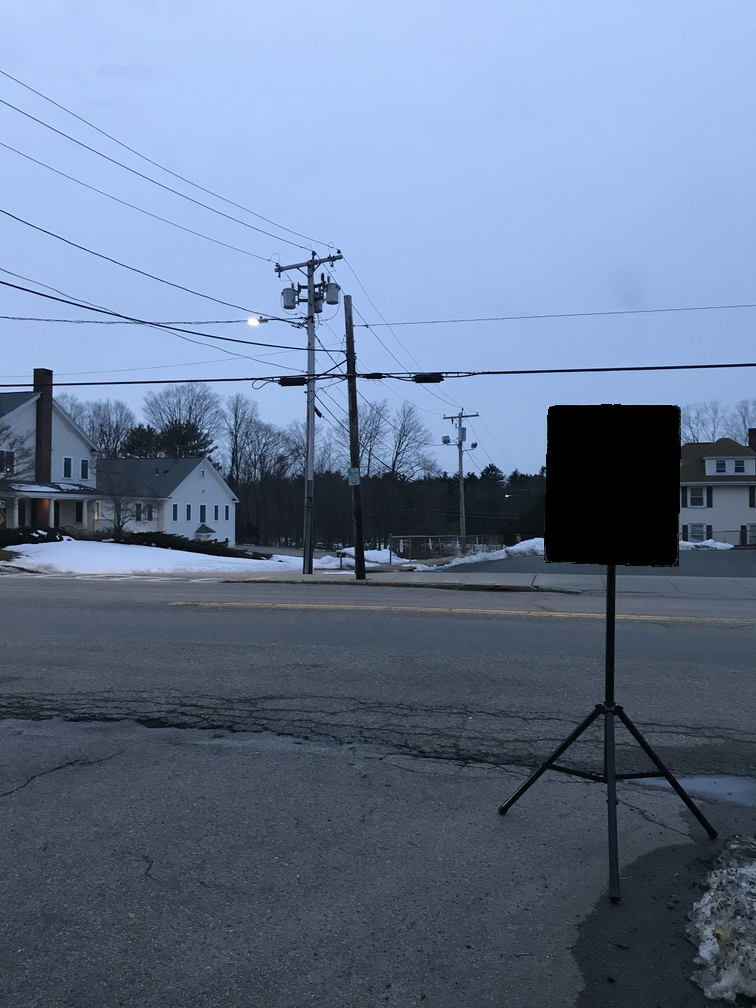}
         \caption{Rectangle}
    \end{subfigure}
    \caption{\textbf{Transfer across patch shapes on the DAPRICOT object detection dataset.}}
    \label{fig:dapricot}
\end{figure}

\subsection{Discussion}
\textbf{Effectiveness of the Patch Detector:} To figure out how the patch detector performs in identifying the corrupted pixels, we conduct quantitative evaluations on the RESISC-45 dataset. Attacks are generated by Masked AutoPGD under the DO and BPDA attack modes. We report the precision, recall, accuracy, and F1 of the adversarial pixel segmentation task on attack images. According to Table~\ref{tab:psp-acc}, our patch detector can effectively identify manipulated pixels under both attack modes, although there is a 1\% drop in Recall from DO attack to the stronger BPDA attack. Empirically, we observe that these 1\% uncovered pixels, especially at the patch border, can have some influence on the overall accuracy. We also evaluated our patch detector on the benign images. The false positive detection rate is 5.05e-06.

\begin{table}[!t]
    \centering
    \begin{tabular}{@{}lcccc@{}}
        \toprule
         & \textbf{Recall}     & \textbf{Prec}  & \textbf{Acc} & \textbf{F1} \\ \cmidrule{1-5}
        \textbf{DO} & 99.8\% &  99.1\% & 99.9\% & 99.5\% \\
        \textbf{BPDA} & 98.8\% & 99.1\% & 99.8\% & 99.0\%\\
        \bottomrule
    \end{tabular}
    \caption{\textbf{Adversarial pixel segmentation performance of the patch detector on RESISC-45.}}
    \label{tab:psp-acc}
\end{table}

\begin{table}[!t]
    \centering
    \begin{tabular}{@{}lccc@{}}
    \toprule
                  & \textbf{MPGD}    & \textbf{MAPGD}    & \textbf{MCW}      \\ \midrule
    \textbf{MPGD} & \textbf{81.07\%} & 80.18\%          & 71.67\%          \\
    \textbf{MAPGD} & \textbf{81.07\%} & \textbf{81.13\%} & 66.97\%          \\
    \textbf{MCW}   & 80.71\%          & 80.72\%          & \textbf{77.41\%} \\ \bottomrule
    \end{tabular}
    \caption{\textbf{Generalization to different attacks on ImageNet.} Each row and column represents a model trained with and tested on a specific type of attack.}
    \label{tab:crossattacks}
\end{table}

\begin{table}[!t]
\centering
    \begin{tabular}{@{}lccc@{}}
        \toprule \textbf{Model} & \textbf{Time} & \textbf{GPU} & \textbf{Param} \\ \cmidrule{1-4}
        \textbf{ResNet-50}     & 8 mins & 1.96 GB   &  25.5M    \\
        \textbf{PC~\cite{xiang2021patchcleanser}}     & 758 mins & 7.32 GB   &    25.5M  \\
        \textbf{PZ}     & 12 mins & 3.33 GB   &  72.2M
        \\ \bottomrule 
    \end{tabular}
    \caption{\textbf{Memory cost and speed.} Inference time is on the entire ImageNet validation dataset with one Nvidia 2080Ti.}
    \label{tab:speed}
\end{table}

\textbf{Transfer across Patch Shapes:}
The patch detector in PatchZero operates at the pixel-level, so it can generalize well to different patch shapes.  We evaluate a version of PatchZero, trained using only square patches, on the Dynamic APRICOT dataset\footnote{https://armory.readthedocs.io/en/latest/scenarios/\#dapricot-object-detection-updated-july-2021}, which contains a mixture of diamond, octagon, and rectangle shape patches. We use FasterRCNN as the downstream detector and the Masked PGD attack with 100 iterations. As shown in Figure~\ref{fig:dapricot},  PatchZero can detect and remove adversarial pixels accurately across the three different shapes. Quantitatively, the mAP of the undefended baseline  drops from 27.33\% to 0\%; in contrast, PatchZero has the same benign accuracy of 27.33\% and maintains 20.67\% mAP after attack.
\label{sec:dis}

\begin{figure}[!t]
    \centering
    \begin{subfigure}[b]{0.31\columnwidth}
         \centering
         \includegraphics[width=\columnwidth, height=0.75\columnwidth]{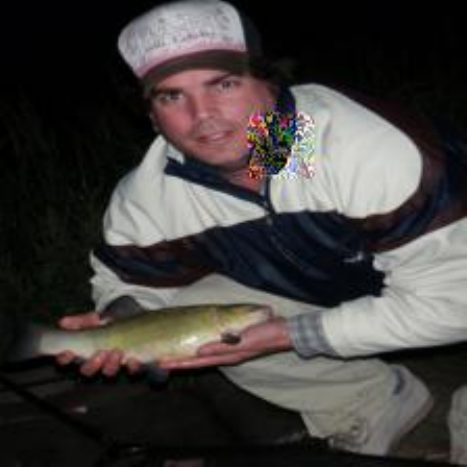}
     \end{subfigure}
     \hspace{0.2em}
     \begin{subfigure}[b]{0.31\columnwidth}
         \centering
         \includegraphics[width=\columnwidth, height=0.75\columnwidth]{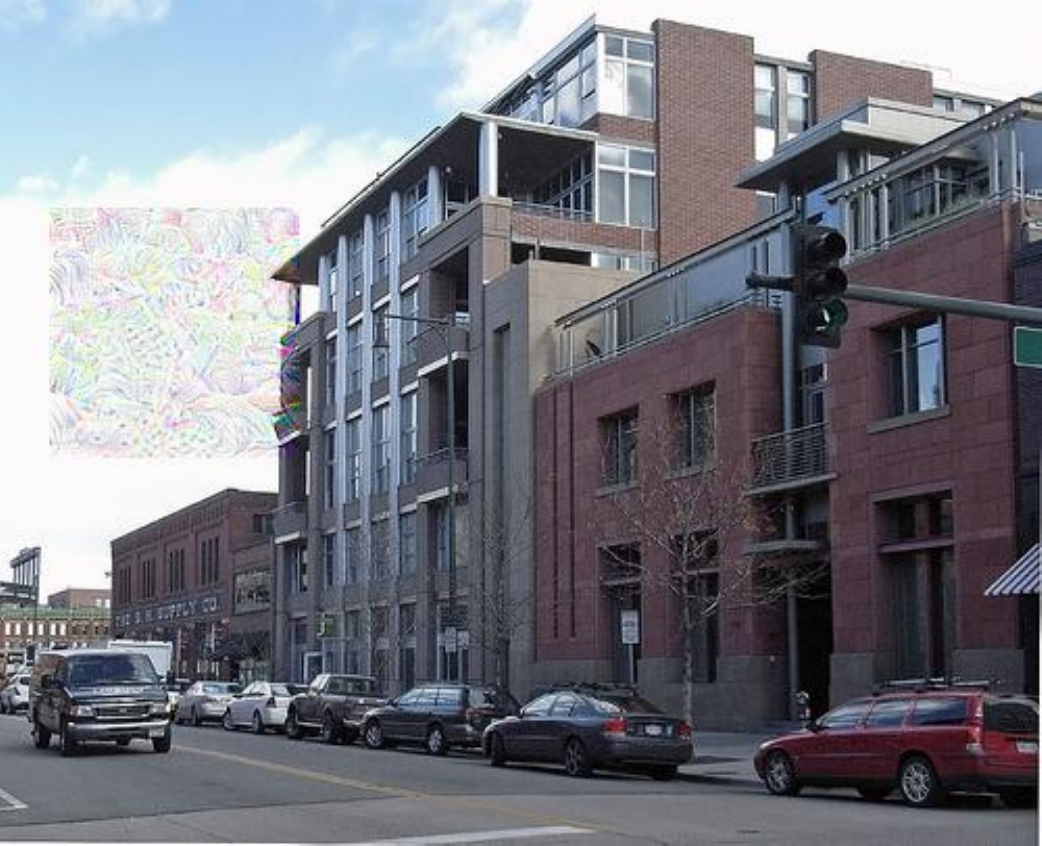}
     \end{subfigure}
    \hspace{0.2em}
    \begin{subfigure}[b]{0.31\columnwidth}
         \centering
         \includegraphics[width=\columnwidth, height=0.75\columnwidth]{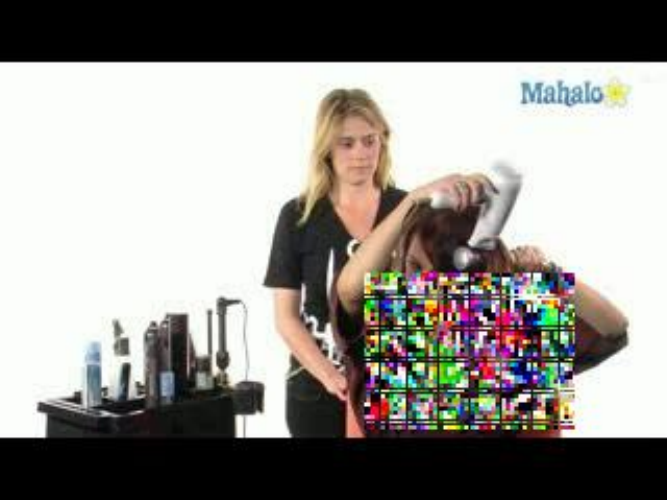}
     \end{subfigure}
     
     \vspace{0.5em}

     \begin{subfigure}[b]{0.31\columnwidth}
         \centering
         \includegraphics[width=\columnwidth, height=0.75\columnwidth]{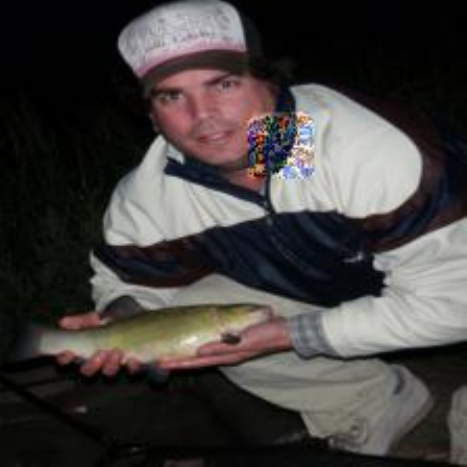}
     \end{subfigure}
     \hspace{0.2em}
     \begin{subfigure}[b]{0.31\columnwidth}
         \centering
         \includegraphics[width=\columnwidth, height=0.75\columnwidth]{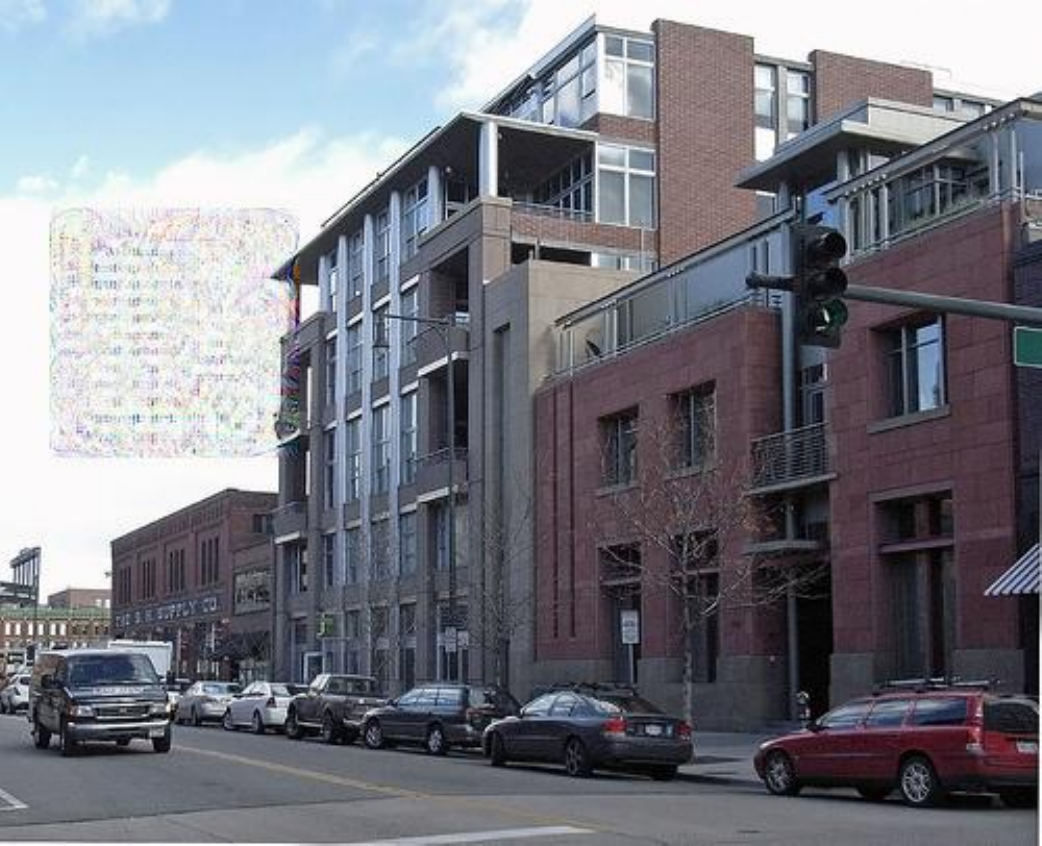}
     \end{subfigure}
     \hspace{0.2em}
     \begin{subfigure}[b]{0.31\columnwidth}
         \centering
         \includegraphics[width=\columnwidth, height=0.75\columnwidth]{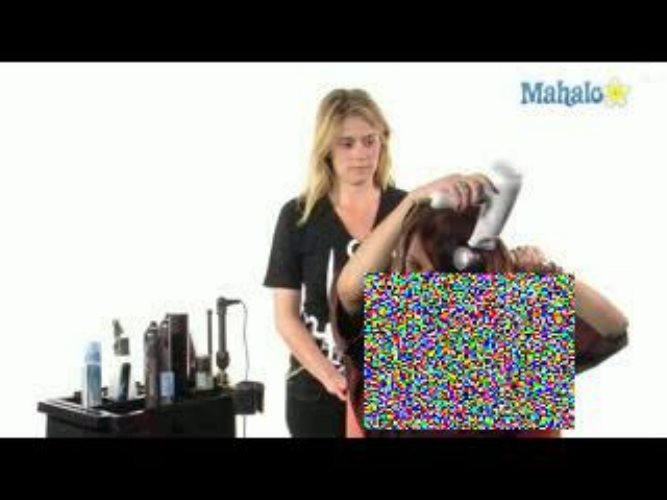}
     \end{subfigure}
        \caption{\textbf{Attack visualization.} We compare the MAPGD attack patterns under the DO (top) and BPDA adaptive attacks (bottom) in the image classification (left), object detection (middle), and video classification tasks (right).}
        \label{fig:pattern}
\end{figure}

\textbf{Transfer across Attack Types:} We performed experiments of training PatchZero on one of the three attacks (MPGD, MAPGD, and MCW) and evaluating the defense performance on all three; results are shown in Table~\ref{tab:crossattacks}. The attacks are DO only on the Imagenet classification task (the numbers are a bit different than those reported in Table 1 of the submitted paper as those results were the BPDA trained models). The results show that, in general, PatchZero trained on one type of attack defends against another attack quite effectively, with only small drops compared to training on the seen attacks. This is particularly the case for the model trained with the MCW attack.

 It is noteworthy that MCW attack is the weakest of the three (it drops the undefended accuracy less) but is harder to defend against and provides the best generalization. A likely explanation for this behavior is that the MPGD and MAPGD patches have high pixel values and produce more distinct patterns from natural images than the MCW patches. A detector that can detect subtle patches generalizes to more distinct patches. 

\textbf{Computation Overhead:} We analyze the memory cost and inference speed of PatchZero in Table~\ref{tab:speed}. Both PatchZero and PatchCleanser use {ResNet50} as backbone. Although Patchzero has more model parameters, it has a faster (60x) inference speed and lower (2x) GPU memory.

\textbf{DO \vs BPDA Attack Patterns:} We compare the DO (top) and BPDA adaptive attack patterns (bottom) in Figure~\ref{fig:pattern}. The left column shows adversarial patches in the image classification task. The BPDA attack pattern seems to be more colorful and granular than the DO attack pattern. For the object detection task (middle), BPDA attack patterns are more structured, rather than a seemingly random appearance. For the video classification task (right), the DO patch shows some ``grid-like'' pattern, while the BPDA patch is denser and more granular. 

In all cases, the two types of attack patterns are very different. The BPDA attack patterns start with the DO patterns and gradually evolve into the granular patterns. The changing appearance requires the patch detector to be updated at each iteration. It is not at all obvious that the process should converge but, thanks to the two-stage training, the trained patch detector becomes robust against the BPDA patches.

\begin{figure}[!t]
    \centering
    \captionsetup{type=figure}
    \begin{subfigure}[b]{0.31\columnwidth}
         \centering
         \includegraphics[width=\textwidth, height=0.75\textwidth]{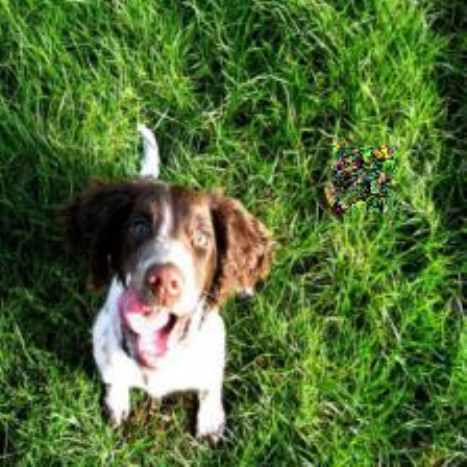}
         \caption{\scriptsize{\textcolor{green}{\textsc{Dog}}, \textcolor{red}{\textsc{Car}}}}
     \end{subfigure}
     \hspace{0.2em}
     \begin{subfigure}[b]{0.31\columnwidth}
         \centering
         \includegraphics[width=\textwidth, height=0.75\textwidth]{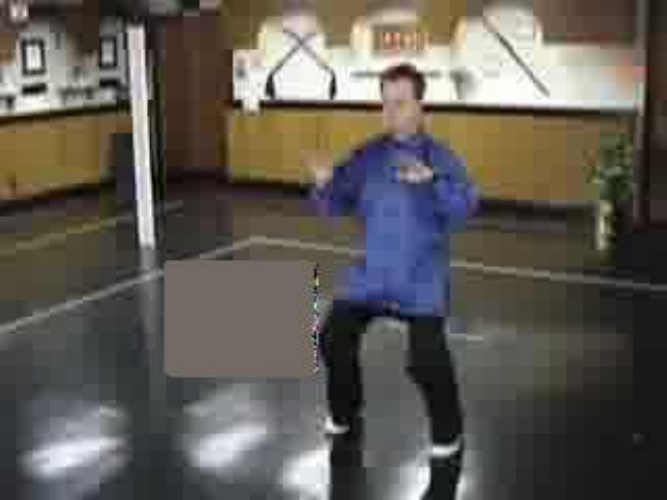}
         \caption{\scriptsize{\textcolor{green}{\textsc{TaiChi}}, \textcolor{red}{ \textsc{YoYo}}}}
     \end{subfigure}
     \hspace{0.2em}
     \begin{subfigure}[b]{0.31\columnwidth}
         \centering
         \includegraphics[width=\textwidth, height=0.75\textwidth]{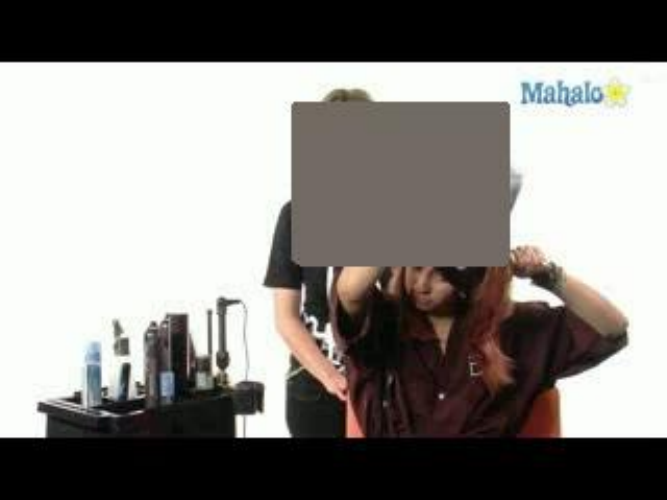}
         \caption{\tiny{\textcolor{green}{\textsc{BlowDryHair}}, \textcolor{red}{\textsc{Haircut}}}}
     \end{subfigure}
    \caption{\textbf{Three common failure cases of PatchZero.} (a) miss patch detection, (b) leaking adversarial pixels, (c) occlusion. Green denotes ground truth labels and red denotes incorrect predictions. Figure best viewed when zoomed in.}
    \label{fig:failcases}
\end{figure}

\textbf{Failure Cases and Limitations:} We present three common failure cases of PatchZero in Figure~\ref{fig:failcases}. As shown in part (a), missed patch detection can lead to defense failure, since repainting cannot be effectively applied without a correct patch detection. In the example shown, the adversarial patch has a similar texture as the background, leading to a missed detection. Leaking adversarial pixels in (b) is another failure case though morphological operations applied to the binary mask prediction can reduce the effects. Final failure cases in part (c) arises due to significant occlusion caused by random patch location falling on top of the main object in the scene, regardless of correct patch detection.

\section{Conclusions}
\label{sec:con}

In this paper, we proposed PatchZero, a general defense pipeline against white-box patch attacks. PatchZero first detects the adversarial pixels and then ``zeros out'' the patch region by repainting with mean pixel values. We further propose a two-stage training scheme to defend against the stronger adaptive attacks. Extensive experiments demonstrate the state-of-the-art robustness of PatchZero across the tasks of image classification, object detection, and video classification, with little degradation in benign performance. PatchZero transfers well to different patch shapes and attack types.


\section*{Acknowledgment}
This research was funded, in part, by the US Government (DARPA GARD LR2 project HR00112020009). The views, opinions, and/or findings expressed are those of the authors and should not be interpreted as representing the official views or policies of the Department of Defense or the U.S. Government.

\clearpage
{\small
\bibliographystyle{ieee_fullname}
\bibliography{egbib}
}

\end{document}